\pdfoutput=1
\documentclass[nohyperref]{article}

\usepackage{microtype}
\usepackage{graphicx}
\usepackage{subfigure}
\usepackage{booktabs} 

\usepackage{hyperref}

\usepackage[accepted]{icml2022}

\usepackage{amsmath}
\usepackage{amssymb}
\usepackage{mathtools}
\usepackage{amsthm}

\usepackage[capitalize,noabbrev]{cleveref}

\usepackage{bbm}
\usepackage{mathtools, nccmath}
\usepackage{amssymb}
\usepackage{lmodern}
\usepackage{xparse}
\DeclarePairedDelimiterX{\set}[1]{\{}{\}}{\setargs{#1}}
\NewDocumentCommand{\setargs}{>{\SplitArgument{1}{;}}m}
{\setargsaux#1}
\NewDocumentCommand{\setargsaux}{mm}
{\IfNoValueTF{#2}{#1} {#1\,\delimsize|\,\mathopen{}#2}}
\usepackage{multirow}
\usepackage{enumitem}

\usepackage{xspace}
\newcommand{\name}{\texttt{FedSpace}\xspace}
\newcommand{\fedbuff}{\texttt{FedBuff}\xspace}

\DeclareMathOperator*{\argmax}{arg\,max}

\theoremstyle{plain}

\theoremstyle{definition}

\theoremstyle{remark}

\usepackage[textsize=tiny]{todonotes}

\icmltitlerunning{\name: An Efficient Federated Learning Framework at Satellites and Ground Stations}

\begin{document}

\twocolumn[
\icmltitle{\name: An Efficient Federated Learning Framework at \\Satellites and Ground Stations}




 \begin{icmlauthorlist}
 \vskip 0.10in
 \icmlauthor{Jinhyun So}{usc}
 \icmlauthor{Kevin Hsieh}{msr}
 \icmlauthor{Behnaz Arzani}{msr}
 \icmlauthor{Shadi Noghabi}{msr}
 \icmlauthor{Salman Avestimehr}{usc}
 \icmlauthor{Ranveer Chandra}{msr}
 \end{icmlauthorlist}

 \icmlaffiliation{usc}{University of Southern California, Los Angeles, California, USA}
 \icmlaffiliation{msr}{Microsoft Research, Redmond, Washington, USA}

 \icmlcorrespondingauthor{Jinhyun So}{jinhyuns@usc.edu}
 \icmlcorrespondingauthor{Kevin Hsieh}{kevin.hsieh@microsoft.com}


\vskip 0.30in
]



\printAffiliationsAndNotice{}  

\begin{abstract}
%
%
%
Large-scale deployments of low Earth orbit (LEO) satellites collect massive amount of Earth imageries and sensor data, which can empower machine learning (ML) to address global challenges such as real-time disaster navigation and mitigation.
However, it is often infeasible to download all the high-resolution images and train these ML models on the ground because of limited downlink bandwidth, sparse connectivity, and regularization constraints on the imagery resolution.
To address these challenges, we leverage Federated Learning (FL), where ground stations and satellites collaboratively train a global ML model without sharing the captured images on the satellites. 
We show fundamental challenges in applying existing FL algorithms among satellites and ground stations, and we formulate an optimization problem which captures a unique trade-off between staleness and idleness.
We propose a novel FL framework, named \name, which dynamically schedules model aggregation based on the deterministic and time-varying connectivity according to satellite orbits.
Extensive numerical evaluations based on real-world satellite images and satellite networks show that \name reduces the training time by 1.7 days (38.6\%) over the state-of-the-art FL algorithms.
%
%
%
%
%
\vspace{-0.2cm}
\end{abstract}

\section{Introduction}\label{sec:intro}
\vspace{-0.1cm}
Advancements in satellite technology has drastically lowered the costs of satellite deployments, which stimulate massive growth in low Earth orbit (LEO) satellites. Multiple companies committed to deploy thousands of small satellites in the next few years~\cite{harris2018techgiant, platnetlabfactory, worldvu, spacex}. These large constellations of satellites collect near real-time satellite imagery and sensor data, which can empower machine learning (ML) to address emerging global challenges, such as food security~\cite{AragonHTFM18}, disaster navigation~\cite{BarmpoutisPDG20, chencascading2020}, climate change~\cite{shuklafsfs21}, and disease spreads~\cite{FRANCHPARDO2020140033}.

The challenge is in these satellite-based applications' ability to train their ML models effectively and in a timely manner. While the nature of these applications makes it necessary to have an accurate (and therefore, up-to-date) model at all times, it is difficult to continuously update these models. 

Training ML models on the ground is increasingly infeasible because we cannot download all the data from the satellites: (a) unlike geostationary (GEO) satellites, LEOs can only communicate with the ground several times a day when they come in contact with a ground station~\cite{denby2020orbital}; and (b) downlink bandwidth is a major bottleneck --- ground stations cost millions of dollars and are difficult to scale~\cite{l2d2}. The massive deployment of satellite constellations with high-resolution cameras further exacerbates this problem as more satellites and more data compete for the limited available bandwidth. Furthermore, downloading high-resolution satellite imagery may not always be possible due to regulation restrictions and privacy concerns~\cite{coffer20balancing}.

The alternative i.e., training the models in a distributed fashion in space, is also currently infeasible: inter-satellite communication is currently not possible in LEO satellites~\cite{Handley19}. Even if satellites could communicate, the amount of memory and power required to do so would quickly become the bottleneck.

Our goal in this work is to design an effective distributed ML framework at satellites and ground stations \emph{without} downloading satellite data to the ground. We leverage Federated Learning (FL)~\cite{KonecnyMYRSB16, McMahanMRHA17}, where servers (ground stations) and clients (satellites) collaboratively train ML models without sharing training data. 

However, it is fundamentally challenging to apply existing FL algorithms to this environment because the connectivity of satellites is sparse and heterogeneous (Section~\ref{subsec:connectivity_sets}). 
This makes existing FL algorithm extremely slow as most algorithms (e.g.,~\citealt{McMahanMRHA17, BonawitzIKMMPRS17, BonawitzEGHIIKK19, KairouzMABBBBCC21}) assume synchronous updates among clients in each communication round, and satellites with limited connectivity become stragglers that make other satellites idle. 
On the other hand, asynchronous FL algorithms~\cite{xie2019asynchronous, van2020asynchronous} introduce large staleness in local updates that lead to model performance degradation. Our evaluation with real-world satellite imagery~\cite{christie2018functional} and constellation~\cite{safyan2020planet} shows synchronous FL is unacceptably slow as it take 45.8 days to train an ML model, while asynchronous FL fails to achieve the target accuracy due to large staleness.

We present \name, an efficient FL framework running among satellites and ground stations. We formulate an optimization problem that maximizes the model convergence rate according to the global model aggregation schedule at the ground stations, which captures the unique trade-offs between satellite \emph{idleness} and local model \emph{staleness} in this environment. To solve this optimization problem, \name first uses satellite orbits and Earth's rotation to calculate the deterministic and time-varying satellite connectivity. \name then uses this information to determine the global model aggregation schedule that makes better trade-offs between idleness and staleness. 

We evaluate \name with one of PlanetLab's satellite networks that consists of 12 ground stations and 191 satellites~\cite{foster2018constellation, safyan2020planet}, and we run our experiments using a satellite simulator~\cite{denby2020orbital} and a real-world satellite dataset~\cite{christie2018functional} over IID and Non-IID dataset distributions. Our evaluation shows \name significantly reduces the training time over synchronous FL~\cite{McMahanMRHA17} and the state-of-the-art buffered asynchronous FL~\cite{nguyen2021federated} by 43.1 days ($16.5\times$) and 1.7 days ($1.7\times$), respectively, all the while achieving the same accuracy.
We make the following contributions:

\begin{itemize}[noitemsep,topsep=0pt,parsep=1pt,partopsep=2pt,leftmargin=*]
    \item We identify the fundamental challenges in applying federated learning at satellites and ground stations for training ML models, as well as the trade-off between satellite idleness and local update staleness.
    
    \item We formulate an optimization problem to maximize the model convergence rate based on the model aggregation schedules at ground stations.

    \item We propose a novel FL framework that solves the optimization problem by leveraging the deterministic and time-varying satellite connectivity.
    
    \item We empirically demonstrate the effectiveness of our solution using real-world satellite networks and satellite imagery dataset, and we show that our solution significantly reduces model training time over the state-of-the-art FL algorithms.
    
\end{itemize}

\section{Background and Motivation}
\label{sec:system}


    

In this section, we introduce the system and connectivity model for FL at the satellites and ground stations. We then discuss why existing FL algorithms fall short for a set of ML applications in space.


\subsection{System Model for FL at Satellites}\label{subsec:system_model}

\begin{figure}[t!]
	\centering
	\includegraphics[height=2.2in, trim=0.05cm 0.05cm 0.1cm 0.1cm, clip]{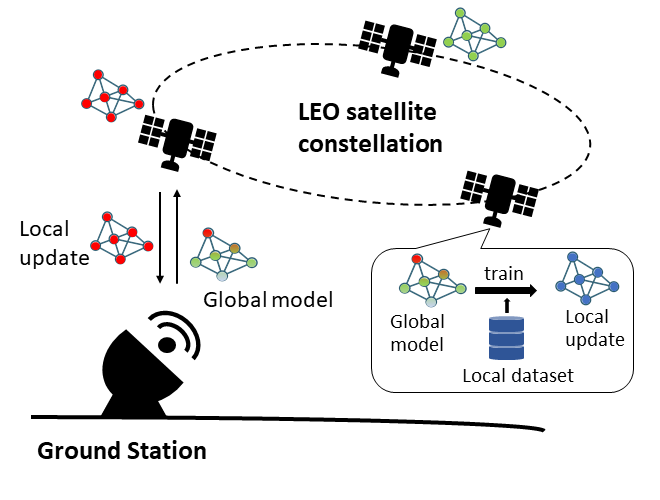}
	\vspace{-0.7cm}\caption{\footnotesize 
    FL framework at satellite networks.
	}
	\label{fig:system}
	\vspace{-0.4cm}
\end{figure}

We consider a constellation of $K$ satellites and $G$ ground stations. 
Satellite $k\in\mathcal{K}=\{1,\ldots,K\}$ collects and stores dataset $\mathcal{D}_k$. The satellites then collaboratively learn a global model $\mathbf{w}\in \mathbb{R}^d$ by minimizing a global objective function as follows

\begin{equation}\label{eq:glob_obj}
    \min_{\mathbf{w}\in \mathbb{R}^d} \Big\{ f(\mathbf{w}) = \sum_{k\in\mathcal{K}} \frac{m_k}{m} f_k(\mathbf{w})\Big\},
\end{equation}
where $m_k=| \mathcal{D}_k |$ and $m=\sum_{i=k}^{K} m_k$ is the size of the whole dataset. Local objective function $f_k$ represents the loss function associated with dataset $\mathcal{D}_k$, i.e., $f_k(\mathbf{w})=\frac{1}{m_k}\sum_{\mathbf{x}\in\mathcal{D}_k} l(\mathbf{w};\mathbf{x})$ where $l(\mathbf{w};\mathbf{x})$ is training loss for a data point $\mathbf{x}$ and model parameter $\mathbf{w}$.

\begin{figure*}[t!]
\centering
    \subfigure[ Number of connected satellites ($=|\mathcal{C}_i|$)]
    {\label{fig:connections_vs_iter}
    \includegraphics[width=0.38\textwidth]{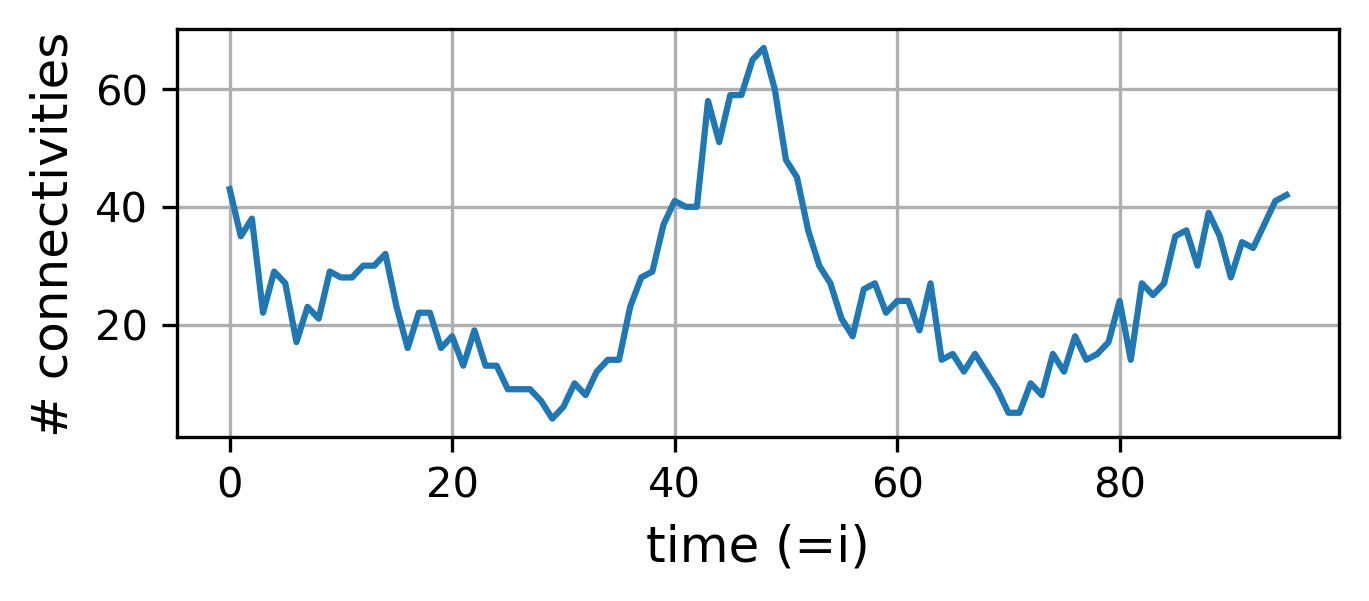}
    }
    \quad\quad
    \subfigure[ Histogram of  $n_k=\sum_{i=0}^{95}\mathbbm{1}\{k\in\mathcal{C}_i\}$, $k\in\mathcal{K}$ ]{\label{fig:hist_connections}
    \includegraphics[width=0.38\textwidth]{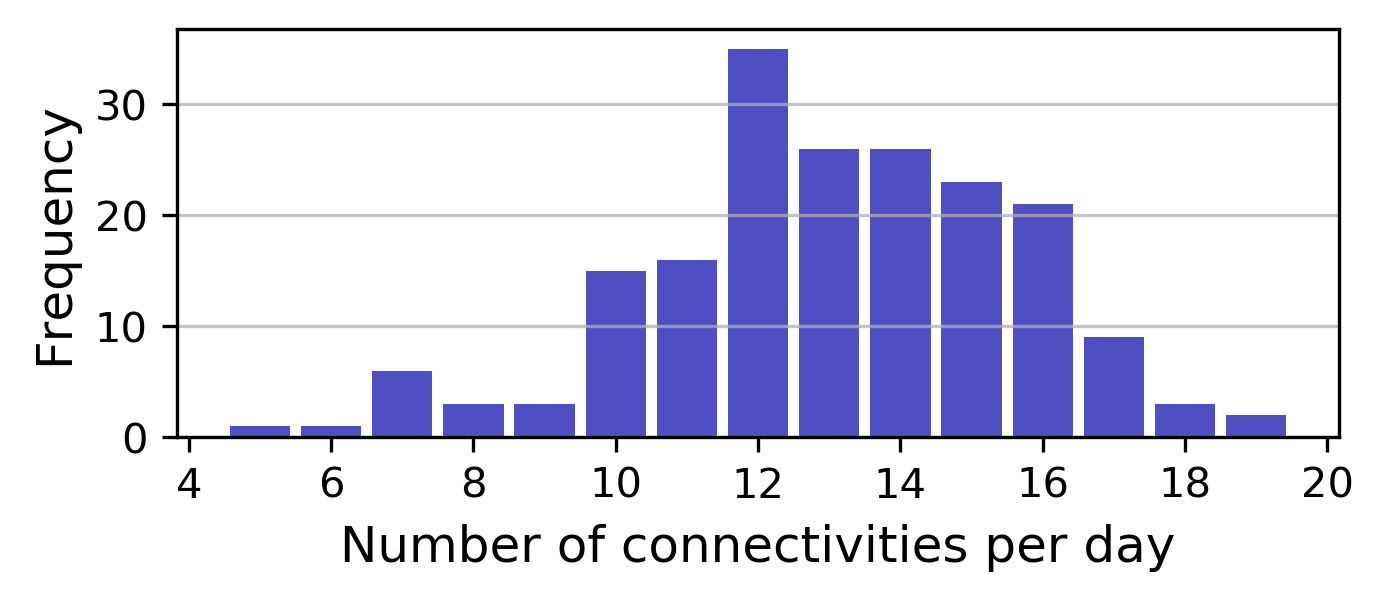}
    }
    \vspace{-4mm}
\caption{Statistics of connectivity sets $\mathcal{C}_i$ defined in \eqref{eq:def_connectivity_set} for a day ($i\in\{0,\ldots,95\}$) with real-world satellite networks consisting of Planet Labs' $12$ ground stations and $191$ satellites~\cite{foster2018constellation, safyan2020planet}.}
\label{fig:connectivity}
\vspace{-3mm}
\end{figure*}

We would be able to solve problem (1) if the ground stations could download the full dataset $\mathcal{D}=\cup_{k\in\mathcal{K}}\mathcal{D}_k$ from the satellites. However, sending all the satellite data to the ground stations is often infeasible (see~\S\ref{sec:intro}).
To address this challenge, we leverage federated learning (FL), where a server coordinates the training process with a set of clients without sharing the training data in the clients~\cite{McMahanMRHA17}. 


We let all the ground stations work as a single FL server (referred as GS) --- the fabric connectivity among ground stations is always well-provisioned and much faster than the satellite to ground links. 
The GS manages the learning process and maintains the current version of the global model $\mathbf{w}$ (Figure~\ref{fig:system}) and sends it to the satellites. It also receives model updates and aggregates the updates into the global model.
On the other hand, the satellites work as the FL clients, which download the global model from the GS, generate local updates using the local dataset, and then send the local updates back to the GS in the next contact.

\subsection{Communication Model and Real-world Satellite Connectivity}\label{subsec:connectivity_sets}

\textbf{Communication Model.} In an earth-centered inertial coordinate system, satellite $k\in\mathcal{K}=\{1,\ldots,K\}$ and ground station $g\in \mathcal{G}=\{K+1,\ldots,K+G\}$ have trajectory $\mathbf{r}_k(t)$ and $\mathbf{r}_g(t)$, respectively, where $t$ is a continuous wall clock time. 
A link between satellite $k$ and ground station $g$ is feasible when satellite $k$ is visible from the ground station $g$ within a minimum elevation angle $\alpha_{\text{min}}$, i.e., $\alpha_{k,g}(t)=\angle\left(\mathbf{r}_g(t),\mathbf{r}_k(t)-\mathbf{r}_g(t)\right) \leq \frac{\pi}{2} - \alpha_{\text{min}}$.
Without loss of generality, we introduce a discrete time index $i\in\{0,1,2,\ldots\}$ and a sequence of connectivity sets $\mathcal{C} =\left\{ \mathcal{C}_{0},\mathcal{C}_{1},\mathcal{C}_{2},\ldots \right\}$.
We say satellite $k$ is \emph{connected to ground station GS} at time index $i$ if a link between satellite $k$ and any ground station is feasible for all $t\in [iT_0, (i+1)T_0)$ where $T_0$ is the wall clock time interval between adjacent time indexes $i$ and $i+1$.
We define the connectivity set $\mathcal{C}_{i}$ as
\begin{align}\label{eq:def_connectivity_set}
    \mathcal{C}_{i} = \{
    k \in \mathcal{K} \;|\; \text{satellite $k$ is connected to GS}\}.
\end{align}

The sequence of connectivity sets $\mathcal{C} =\left\{ \mathcal{C}_{0},\mathcal{C}_{1},\mathcal{C}_{2},\ldots \right\}$ is \emph{time-varying} --- satellites orbit and Earth rotates; and \emph{deterministic} --- the GS can predict the future connectivity based on the location and trajectory of the ground stations and satellites. We assume satellites cannot communicate with each other, which is the case in today's constellations~\cite{Handley19}.

\textbf{Real-world satellite connectivity.} We show the number of satellites connected to ground stations  at any given point in time, $i$, in a given day for an example constellation --- that of Planet Lab with $12$ ground stations and $191$ satellites~\cite{foster2018constellation, safyan2020planet}---  in Figure~\ref{fig:connections_vs_iter}. Here, we use the \texttt{cote} simulator \cite{denby2020orbital} to identify $\mathcal{C}$ with $T_0=15$ minutes. 
We observe two types of heterogeneity in $\mathcal{C}$.
\begin{itemize}[noitemsep,topsep=0pt,parsep=1pt,partopsep=2pt,leftmargin=*]
    \item \textbf{Heterogeneous connectivity over time.} As Figure~\ref{fig:connections_vs_iter} shows, the number of connected satellite $|\mathcal{C}_i|$ changes significantly over time. The maximum and minimum value of $|\mathcal{C}_i|$ are $68$ and $4$, respectively. 
    
    \item \textbf{Heterogeneous connectivity among satellites.} Figure~\ref{fig:hist_connections} shows a histogram of the number of connections per day $n_k$ for satellite $k$. $n_k=\sum_{i=0}^{95}\mathbbm{1}\{k\in\mathcal{C}_i\}$ where $\mathbbm{1}\{\cdot\}$ is an indicator function. We observe large variance in $n_k$ e.g, one satellite has only 5 connections per day while another has 19. 
\end{itemize}

Such heterogeneity makes it challenging to run existing FL algorithms in this environment: satellites with lesser connectivity limit the number of communication rounds per day and render other satellites idle while updating the global model without waiting for these satellite introduces large staleness that degrades accuracy. Section \ref{subsec:exsiting_FL} discusses this trade-off in detail.

\subsection{FL at Satellites and Ground Stations}

\textbf{Global model index.} At time index $i$, GS maintains a global model $\mathbf{w}^{i}$ and global training round index $i_g\in\{0,1,2,\ldots\}$, which is incremented by one \emph{only when} the GS updates the global model.

\textbf{FL process at satellites.} The process at satellites is similar to that in typical FL clients. When satellite $k$ is connected to the GS, i.e., $k\in\mathcal{C}_i$, the pair of $\mathbf{w}^{i}$ and $i_g$ is sent to satellites.
%
Satellite $k$ initializes the local model as $\mathbf{w}^{0}_k = \mathbf{w}^{i_{g,k}}$ and stores the training round index $i_g$ of the \emph{base} model as $i_{g,k}$.
Then satellite $k$ locally trains the model by carrying out SGD steps over its own dataset $\mathcal{D}_k$: 
\begin{equation}\label{eq:local_sgd}
    \mathbf{w}_k^{j+1} = \mathbf{w}^{j}_k - \eta \nabla f(\mathbf{w}^{j}_k;\mathbf{X}_k^j),
\end{equation}
where $j$ is a local training index, $\mathbf{X}_k^j$ is a mini-batch of size $B$ selected from $\mathcal{D}_k$ at $j$, and $f(\mathbf{w},\mathbf{X})=\frac{1}{B}\sum_{\mathbf{x}\in\mathbf{X}}l(\mathbf{w},\mathbf{x})$ is a stochastic gradient with random mini-batch $\mathbf{X}$. We assume $\mathbb{E}_{\mathbf{X}\sim\mathcal{D}_k}[f(\mathbf{w},\mathbf{X})] = f_k(\mathbf{w})$ where $f_k$ is the local objective function of satellite $k$ defined in \eqref{eq:glob_obj}.
After $E\geq1$ steps, satellite $k$ stores the gradient $\mathbf{g}_k = \mathbf{w}_k^E - \mathbf{w}_k^0$ and the training round index of the base global model $i_{g,k}$. 
At the next connection, satellite $k$ uploads the pair of $(\mathbf{g}_k,i_{g,k})$ and the GS stores the pair in the buffer $\mathcal{B}$.

\begin{algorithm}[t!]
  \caption{Ground Stations (GS) Procedure}
   \label{alg:GS}
    \begin{algorithmic}
    \STATE {\bfseries Input:} model $\mathbf{w}^{0}$
    \STATE Initialize $i=i_g=0$, $\mathcal{B}_0 = \emptyset$
    
    \REPEAT 
        \FOR{$k \in \mathcal{C}_i$}
            \STATE Receive $((\mathbf{g}_{k}, i_{g,k}))$ from satellite $k$
            \STATE $\mathcal{B}_i \gets \mathcal{B}_i \cup \{(\mathbf{g}_{k}, s_k)\}$ where $s_k =i_g-i_{g,k}$
            \STATE $\mathcal{R}_i \gets \mathcal{R}_i \cup \{k\}$
        \ENDFOR
        
        \STATE $a^{i} = \textsc{Scheduler}(\mathcal{C}_i, \mathcal{B}_i, \mathcal{R}_i) \in \{0,1\}$
        \IF{$a^{i} = 1$} 
            \STATE $\mathbf{w}^{i+1} \gets \textsc{ServerUpdate}\left( \mathbf{w}^{i}, \mathcal{B}_i \right)$
            \STATE $i_g \gets i_g +1$; $\mathcal{B}_{i+1} \gets \emptyset$; $\mathcal{R}_{i+1} \gets \emptyset$
        \ELSE
            \STATE $\mathbf{w}^{i+1} \gets \mathbf{w}^{i}$; $\mathcal{B}_{i+1} \gets \mathcal{B}_{i}$; $\mathcal{R}_{i+1} \gets \mathcal{R}_{i}$
        \ENDIF

        \STATE Broadcasts $(\mathbf{w}^{i+1}, i_g)$ to satellites in $\mathcal{C}_i$
        \STATE $i \gets i+1$
    \UNTIL{stopping criterion is met}
    \end{algorithmic}
\end{algorithm}

\textbf{FL process at the GS.} The GS updates the global model using the model updates stored in the buffer $\mathcal{B}_i$. At each time index $i$, the GS determines whether it should update the global model or not according to its model aggregation algorithm (Section \ref{subsec:exsiting_FL}):
\begin{equation}\label{eq:update_eq_global}
    \mathbf{w}^{i+1} =
    \left\{
    \begin{array}{ll}
        \mathbf{w}^{i} + \sum_{(i_{g,k},\mathbf{g}_k)\in\mathcal{B}_i} \frac{c(s_k)}{C}\mathbf{g}_k, \text{ if } a^i=1 \\
        \mathbf{w}^{i},\;\quad\quad\qquad\qquad\qquad\qquad\text{if } a^i=0
    \end{array} 
    \right. 
\end{equation}
where $a^i$ is an aggregation indicator at time stamp $i$ ($a^i=1$ indicates global model aggregation), and $s_k=i_g - i_{g,k}$ denotes the \emph{staleness} of gradient $\mathbf{g}_k$, $C=\sum_{i_{g,k} \in \mathcal{B}_i}c(s_k)$.
$c(s)$ is a staleness compensation function satisfying $c(0)=1$ and is monotonically decreasing as $s$ increases~\cite{xie2019asynchronous}. We use a polynomial function $c_{\alpha}(s) = (s +1)^{-\alpha}$ in our experiments as it shows similar or better performance than the other options.
%
Once updated, the GS sends the updated global model to the connected satellites ($\mathcal{C}_{i}$).
Algorithm \ref{alg:GS} describes the procedures of GS.


\subsection{Existing FL algorithms} \label{subsec:exsiting_FL}

\begin{figure*}[t!]
\centering
    \subfigure[ Synchronous FL ]{\label{fig:example_sync}
    \includegraphics[height=2.4in, trim=0.05cm 0.05cm 0.1cm 0.1cm, clip]{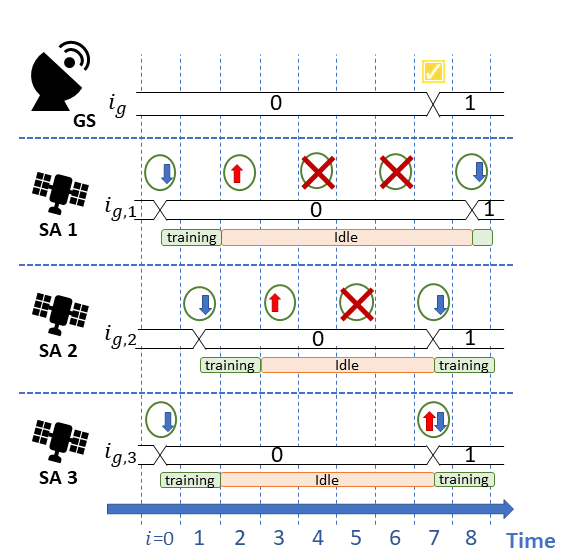}
    }
    \subfigure[ Asynchronous FL ]{\label{fig:example_async}
    \includegraphics[height=2.4in, trim=0.05cm 0.05cm 0.1cm 0.1cm, clip]{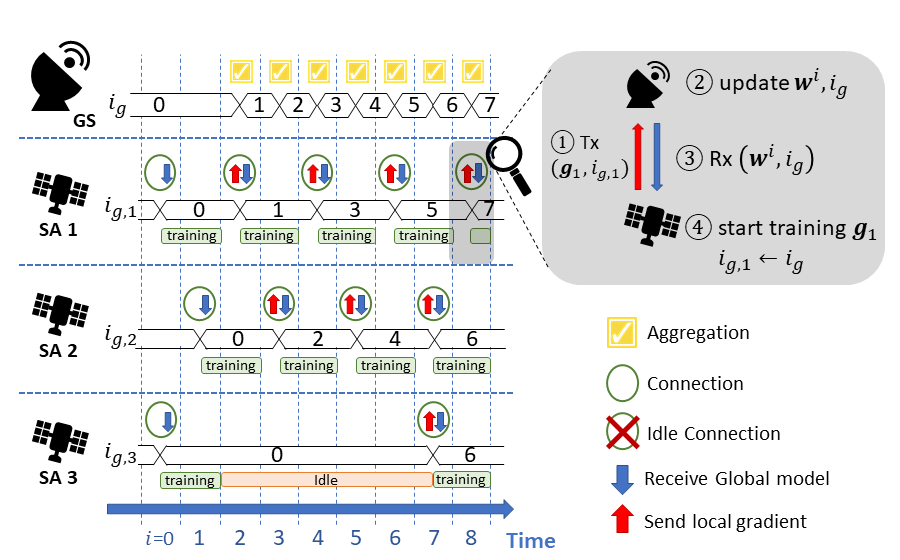}
    }
    \vspace{-4mm}
\caption{The timing diagram of (a) synchronous FL and (b) asynchronous FL in an illustrative example. In (a), all downloaded local gradients have zero staleness, i.e., global training index $i_g$ of GS and $i_{g,k}$ (training index of based model at satellite $k$) are the same, but the satellite with limited connectivity (SA 3) works as a \emph{straggler}. In (b), there is no idle connection, but downloaded local gradients have non-zero staleness. Notably, staleness of the third satellite at $i=7$ is $i_g- i_{g,3}=5$, which can severely degrade the global model. 
Appendix~\ref{app:example} explains this example in detail.
}
\label{fig:example_1}
\vspace{-3mm}
\end{figure*}

\noindent\textbf{Synchronous FL.}
The vast majority of existing FL algorithms assumes synchronous FL (e.g.,~\citealt{McMahanMRHA17, BonawitzIKMMPRS17, BonawitzEGHIIKK19, KairouzMABBBBCC21}), where all local updates are based on the same global model. In synchronous FL, the GS waits for the local gradients from all the satellites before updating the global model.
The indicator variable $a^i$ in \eqref{eq:update_eq_global} is: 
\begin{equation}\label{eq:schdl_syncFL}
    a^{i}_{\text{sync}} = \mathbbm{1}\{\mathcal{R}_i=\mathcal{K}\},
\end{equation}
where $\mathbbm{1}\{\cdot\}$ is an indicator function, $\mathcal{R}_i$ is a index set of satellites whose local gradients are stored in the buffer, and $\mathcal{K}$ is the index set of all satellites. 
Figure~\ref{fig:example_sync} illustrates the timing diagram of synchronous FL using an example with three satellites. The satellite with limited connectivity (SA 3) becomes the straggler that leads to idle connectivities at SA 1 and SA 2.

\noindent\textbf{Asynchronous FL.}
In asynchronous FL~\cite{xie2019asynchronous, van2020asynchronous}, the GS updates the global model whenever local gradients are available.
Hence, the indicator variable $a^i$ is:
\begin{equation}\label{eq:schdl_asyncFL}
    a^{i}_{\text{async}} = \mathbbm{1}\{\mathcal{R}_i\neq\emptyset\}.
\end{equation}
Figure~\ref{fig:example_async} illustrates the timing diagram of the same example. While there is no idle connection in asynchronous FL, there is a large staleness when SA 3 uploads its local gradients at time index $i=7$. Specifically, the local gradient sent from SA 3 at $i=7$ is outdated as the global model has been updated 5 times. Local update with large staleness can negatively impact model training even with staleness compensation.

\noindent\textbf{Buffered asynchronous FL.}
\fedbuff~\cite{nguyen2021federated} is designed to balance between synchronous and asynchronous FL.
The GS stores the local gradients from satellites in a \emph{buffer}, and the GS updates the global model only when the size of the buffer reaches a threshold $M$. 
The indicator variable $a^i$ of \fedbuff is:
\begin{equation}\label{eq:schdl_fedbuff}
    a^{i}_{\text{fedbuff}} = \mathbbm{1}\{|\mathcal{R}_i|\geq M\}.
\end{equation}

\begin{figure}[t!]
	\centering
	\vspace{-0.3cm}
	\includegraphics[height=2.6in, trim=0.05cm 0.05cm 0.1cm 0.1cm, clip]{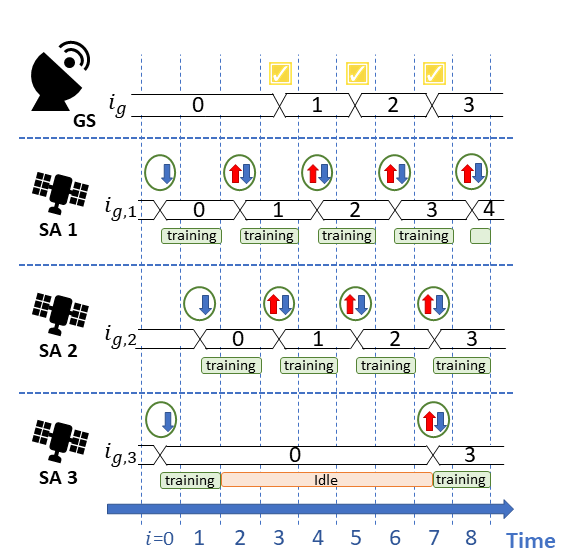}
	\vspace{-0.4cm}\caption{\footnotesize The timing diagram of \fedbuff with buffer size $M=2$. 
	In \fedbuff, there is no idle connection and the largest value of staleness is reduced from $5$ to $2$ when compared to the asynchronous FL. 
	}
	\label{fig:example_fedbuff}
	\vspace{-0.4cm}
\end{figure}

Figure~\ref{fig:example_fedbuff} shows the timing diagram of \fedbuff with $M=2$. Compared to asynchronous FL, \fedbuff also has no idle connections, and the staleness of SA 3 is reduced from 5 to 2. 


\textbf{Summary.} Table~\ref{tbl:summary_example} summarizes the trade-offs between idleness and staleness when applying existing FL algorithms to the illustrative example (Figures \ref{fig:example_1} and \ref{fig:example_fedbuff}). 
We observe that frequent aggregation (asynchronous FL) results in reduction of the number of idle connections but degrades the quality of the local gradient due to larger staleness. 
On the other hand, sparse aggregation (synchronous FL) improves the quality of local gradients by reducing staleness, but decreases the number of aggregated local gradients due to idleness. 
\fedbuff appears to balance between idleness and staleness, but it is unclear if it makes the right trade-offs.

\begin{table}[h]
\centering
\footnotesize 
\vspace{-0.3cm}
\caption{Summary of three FL algorithms in the illustrative examples in Figure~\ref{fig:example_1} and Figure~\ref{fig:example_fedbuff}. $s$ denotes staleness of local gradients to be aggregated to update the global model. ``Idle'' means the case where a satellite is connected to GS but does not send local gradient as it has no update after the previous visit.}
\label{tbl:summary_example}
\scalebox{0.95}{
\begin{tabular}[b]{|c|c|c|c|c|c|c|c|} 
    \hline
    \multirow{2}{*}{Scheme} & {$\#$ global} & \multicolumn{6}{c|}{$\#$ aggregated local gradients}  \\
    \cline{3-8}
       & updates & $s=0$ & 1 & 2 & 5 & Total & Idle\\
    \hline
    Sync      & 1 & 3 & - & - & - & 3 & 5\\
    \hline
    Async     & 7 & 4 & 3 & - & 1 & 8 & 0\\
    \hline
    \fedbuff  & 3 & 7 & - & 1 & - & 8 & 0\\
    \hline
\end{tabular}
}
\vspace{-0.5cm}
\end{table}

\section{The \name Framework}
\label{sec:proposed}

    
    

Before introducing key intuition and procedure of \name, we first define aggregation scheduling vector, $s$-staleness, idleness, and staleness vector, which will be used to formulate an optimization problem for aggregation scheduler.

Aggregation scheduling vector at $i$ is defined as
\begin{equation}\label{eq:def:aggr_vector}
    \mathbf{a}^{i,i+I_0} = [a^i,a^{i+1},\ldots,a^{i+I_0-1}]^\top \in \{0,1\}^{I_0},
\end{equation}
where $I_0$ is a scheduling period. 
Our goal is to design $\mathbf{a}^{i,i+I_0}$ at $i\in\{0,I_0,2I_0,\ldots\}$.
We also define a index set $\mathcal{I}_\text{agg}(\mathbf{a}^{i,i+I_0}) = \{ l\in [i, i+I_0) \;|\;a^l=1\}$.

We say local gradient has $s$-staleness when global model is updated $s$ times before the gradient is sent back to GS, i.e., $s$ denotes the difference between training round indexes of the current global model and the base global model.
Therefore, staleness of the local gradient $\mathbf{g}_k$ at $i$ can be expressed as 
\begin{equation}\label{eq:def_staleness}
    s^i_k = \sum_{l=i'_k}^{i-1} \mathbbm{1}\{a_{l}=1\},
\end{equation}
where $i'_k$ is the latest time index when satellite $k$ is connected to GS before $i$.

We say connectivity of satellite $k$ at $i$ is idle if satellite $k$ has no local update at $i$ even though it is connected to the GS. It corresponds to the case that satellite $k$ did not receive the global model at $i'_k$, i.e., there was no aggregation between $i'_k$ and $i''_k$. $i''_k$ is the latest time index when satellite $k$ is connected to GS before $i'_k$ (e.g., see $i=4$, $k=1$ in Figure~\ref{fig:example_sync}).
Therefore, an indicator of idle connectivity of satellite $k$ at $i$ can be expressed as
\begin{equation}\label{eq:def_idleness}
    \text{idle}_k^i = \mathbbm{1}\big\{ \sum_{l=i''_k}^{i'_k-1}a_l = 0 \big\}.
\end{equation}

Given $\mathbf{a}^{i,i+I_0}$, we define staleness vector $\mathbf{s}^l \in \mathbb{N}^{K}$ for $l\in\mathcal{I}_\text{agg}(\mathbf{a}^{i,i+I_0})$ as $k$-th element of $\mathbf{s}^l$ is $s^l_k$ if local update from satellite $k$ is stored in the buffer $\mathcal{B}_l$. 
If not, $k$-th element of $\mathbf{s}^l$ is set as $-1$ which indicates satellite $k$ does not contribute to this global model update. For example, in Figure~\ref{fig:example_sync}, $\mathbf{s}^7=[0,0,0]^\top$ and in Figure~\ref{fig:example_async}, $\mathbf{s}^7=[-1,1,5]^\top$ at $i=7$.

\subsection{Key insight and Optimization Problem}
The key insight of \name is that the connectivity set $\mathcal{C}_i$ is time-varying but deterministic. Given aggregation pattern $\mathbf{a}$, GS can accurately calculate the staleness vector $\mathbf{s}^l$ for all $l\in\mathcal{I}_\text{agg}(\mathbf{a})$.
Therefore, if we know the expected performance gain with respect to model aggregation with certain staleness vector, we can find an optimal aggregation pattern $\mathbf{a}^{i,i+I_0}$ which maximizes the sum of performance gain. 
Our optimization problem can be expressed by
\begin{equation}\label{eq:opt_problem}
    \mathbf{a}^{i,i+I_0}_\text{opt} = \argmax_{\mathbf{a}\in\{0,1\}^{I_0}} \sum_{l\in\mathcal{I}_\text{agg}(\mathbf{a})}u\left(\mathbf{s}_l, \mathcal{T}_l\right),
\end{equation}
where $u$ is a utility function and $\mathcal{T}_l$ denotes training status of global model $\mathbf{w}^l$ at $l$.
The utility $u(\mathbf{s}_l, \mathcal{T}_l)$ is the reduced amount of objective (or loss) function in~\eqref{eq:glob_obj} from model update at $l$ with staleness vector $\mathbf{s}_l$.
We include $\mathcal{T}_l$ as an input of $u$ because the utility may have different value according to training status. For instance, when the global model is almost converged and hence does not change over training round index, local gradient with large staleness does not degrade the utility. On the other hand, at the beginning stage of training, as the global model changes much over each update, local gradient with larger staleness can severely degrades the utility. 
Now, we state how \name solves the optimization problem in \eqref{eq:opt_problem}.

\begin{figure}[t!]
	\centering
	\includegraphics[width=0.48\textwidth]{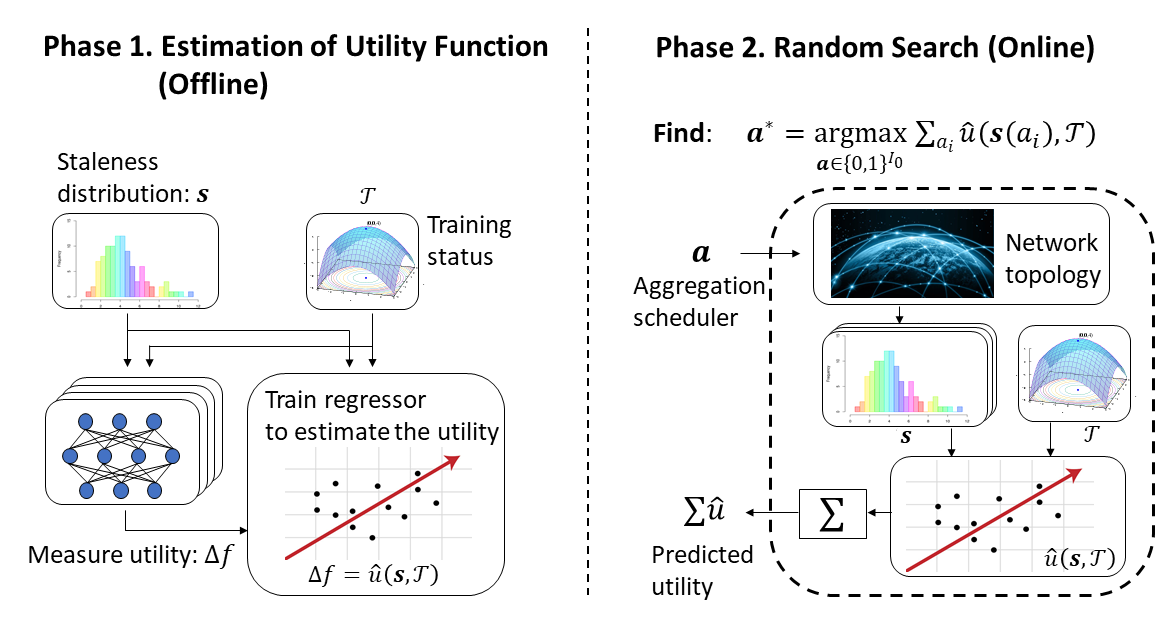}
	\vspace{-0.6cm}\caption{\footnotesize 
    Overview of aggregation scheduler in \name.
	}
	\label{fig:overview}
	\vspace{-0.2cm}
\end{figure}

\subsection{Model Aggregation Scheduler of \name}
Figure~\ref{fig:overview} shows the model aggregation scheduler of \name, which consists of two phases. 
In the first phase, GS estimates the utility function $u$ in \eqref{eq:opt_problem}. To do so, GS generates the pairs of input and output of utility function and train a regression model based on the pairs. 
In the second phase, GS solves the optimization problem approximately by utilizing random search with the regression model trained in the first phase. 

\noindent\textbf{Estimation of Utility Function.}
To generate the pairs of input and output of the utility function, 
the first step is to train ML model parameters with $\mathcal{D}^{\text{s}}$ and stores the sequence of the trained models $\{\mathbf{w}^{i_g}\}_{i_g\in\{0,1,\ldots,I_\text{max}\}}$ where $I_\text{max}$ is the number of total training rounds. 
$\mathcal{D}^{\text{s}}$ is source dataset that has the same task as target dataset $\mathcal{D}=\cup_{k\in\mathcal{K}}\mathcal{D}_k$ defined in \eqref{eq:glob_obj}.
Next, GS randomly generates the input pairs of ($\mathbf{s}$, $i_\text{start}$) from $[-1,0,\ldots,s_\text{max}]^{K}$ and $[0,1,\ldots,I_\text{max}]$.
Then GS measures the amount of reduced loss by applying the pairs into the pretrained ML parameters as
\begin{equation}
    \Delta f = f(\mathbf{w}^{i_\text{start}}) - f_s\left(\mathbf{w}^{i_\text{start}} - \sum_{k=1}^{K}\mathbbm{1}\{s_k\geq0\}\mathbf{g}_k(s_k)\right)
\end{equation}
where $f$ is a objective (or loss) function associated with dataset $\mathcal{D}^{\text{s}}$, $\mathbf{g}_k(s_k)$ is gradient of $f$ with respect to $\mathbf{w}^{i_\text{start} - s_k}$, and $s_k$ is $k$-th element of staleness vector $\mathbf{s}$.
For the training status $\mathcal{T}$ in \eqref{eq:opt_problem}, we use the loss at $i_\text{start}$, i.e., $\mathcal{T}=f(\mathbf{w}^{i_\text{start}})$.
By utilizing $N$ samples of input pair ($\mathbf{s}$, $\mathcal{T}$) and output $\Delta f$, GS trains a regression model $\hat{u}$ such that $\Delta f = \hat{u}(\mathbf{s}, \mathcal{T})$.

\noindent\textbf{Random Search.} Recall that our goal is to find the aggregation scheduling vector $\mathbf{a}^{i,i+I_0}$ defined in \eqref{eq:def:aggr_vector}, which denotes aggregation pattern for next $I_0$ time indexes from $i$. 
Combining~\eqref{eq:opt_problem} and the regression model $\hat{u}$ trained in the first phase, GS finds the best aggregation vector by utilizing random search as
\begin{equation}\label{eq:random_search}
    \mathbf{a}^{i,i+I_0}_\text{*} = \argmax_{\mathbf{a}\in\mathcal{R}} \sum_{l\in\mathcal{I}_\text{agg}(\mathbf{a})}\hat{u}\left(\mathbf{s}_l, f(\mathbf{w}^i)\right)
\end{equation}
where $\mathcal{R}\subset\{0,1\}^{I_0}$ is search domain.
When we set $\mathcal{R}=\{0,1\}^{I_0}$, the search space exponentially increases with respect to $I_0$ which is the scheduling period of \name. 
To reduce the search space, 
we set the range of reasonable number of aggregation, i.e., $n_\text{agg}\in[N_\text{min},\ldots,N_\text{max}]$, that mostly yield positive utility. We infer $N_\text{min}$, and $N_\text{max}$ from $\hat{u}$.
For each trial of $\mathbf{a}$ with $n_\text{agg}$, we randomly select $n_\text{agg}$ positions out of $I_0$ and assign $1$ to the selected positions while assigning $0$ to the other positions in $\mathbf{a}$.
Section \ref{sec:experiments} provides the selection of $I_0$, $N_\text{min}$, and $N_\text{max}$ in our evaluation.

    
    

\section{Experiments}
\label{sec:experiments}
\begin{figure*}[h!]
\centering
    \subfigure[ IID setting ]{\label{fig:accuracy_iid}
    \includegraphics[height=1.6in, trim=0.05cm 0.05cm 0.1cm 0.1cm, clip]{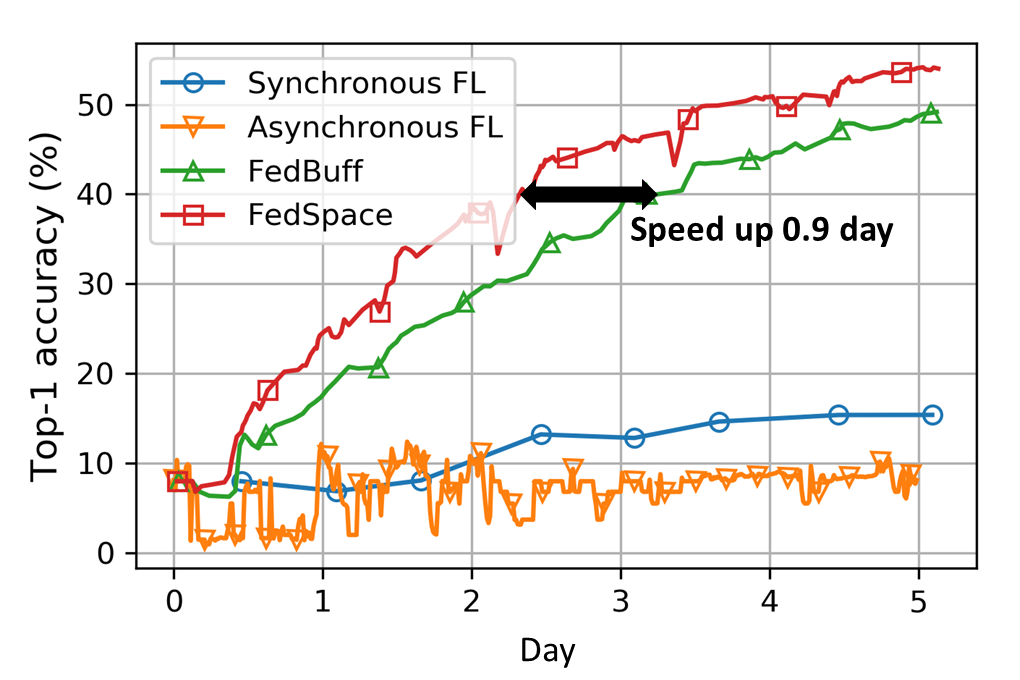}
    }
    \quad\quad
    \subfigure[ Non-IID setting ]{\label{fig:accuracy_noniid}
    \includegraphics[height=1.6in, trim=0.05cm 0.05cm 0.1cm 0.1cm, clip]{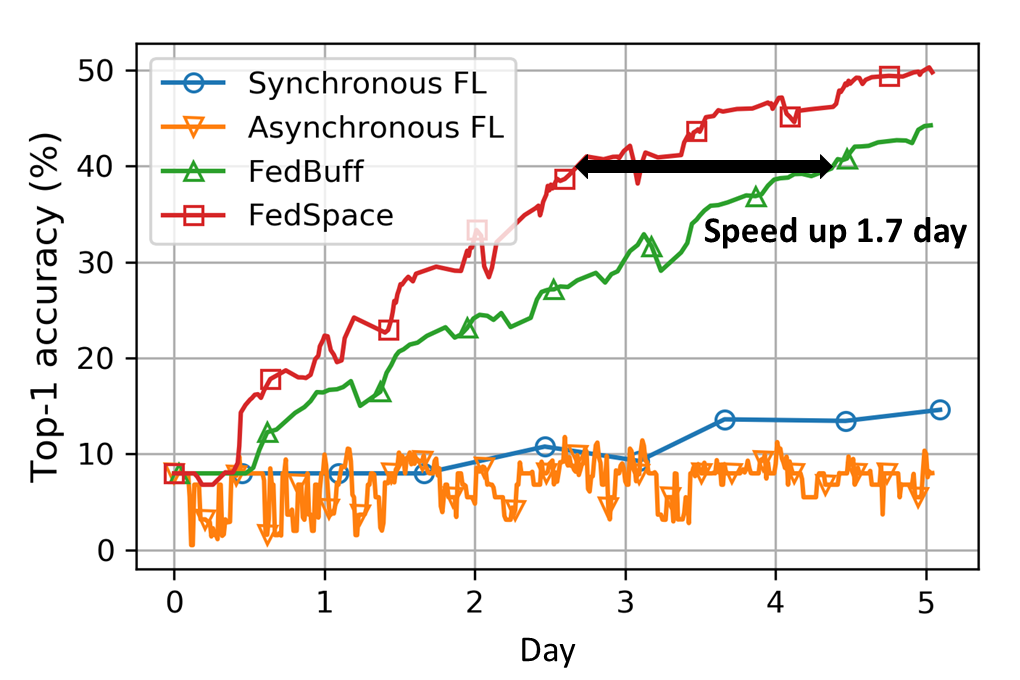}
    }
    \vspace{-4mm}
\caption{Top-1 validation accuracy of DenseNet-161 on fMoW dataset over real-world satellite networks of PlanetLab's $12$ ground stations and $191$ satellites.}
\label{fig:accuracy}
\vspace{-6mm}
\end{figure*}

In this section, we empirically demonstrate that \name significantly outperforms existing FL algorithms in terms of training time to achieve a target test accuracy with a real-world satellite imagery dataset and  a satellite constellation. 

\subsection{Setup}

\noindent\textbf{Satellite Constellation.}
We use the satellite orbits and ground station locations from one of PlanetLab's satellite constellation, which consists of $12$ ground stations and $191$ satellites~\cite{foster2018constellation, safyan2020planet}. We run a satellite constellation simulator to obtain the connectivity information $\mathcal{C}$~\cite{denby2020orbital}.
We set wall clock time period $T_0$ as $15$ minutes between two adjacent time index $i$ and extract $\mathcal{C}$ for $5$ days, i.e., $\mathcal{C}=\{\mathcal{C}_0,\mathcal{C}_1,\ldots,\mathcal{C}_{479}\}$.

\noindent\textbf{Dataset.}
We use the Functional Map of the World (fMoW) dataset, which aims to develop ML models to predict the functional purpose of buildings and land from sequences of satellite images and metadata features~\cite{christie2018functional}.
Each image contains one bounding box with annotated label out of $62$ categories such as construction site, flooded road, educational institution, and etc. 
The metadata provided with each image contains location, time, sun angles and other features to help predictions about the category.

We consider two settings for partitioning the fMoW dataset across the satellites.
\begin{itemize}[noitemsep,topsep=0pt,parsep=1pt,partopsep=2pt,leftmargin=*]
    \item \textbf{IID Setting.} In this setting, the $360,\!000$ training samples are shuffled and partitioned uniformly across the $K=191$ satellites. 
    
    \item \textbf{Non-IID Setting.} In this setting, training samples are assigned to the satellites according to the location of images and trajectory of the satellites. We first partition the training samples according to the UTM zone. For each UTM zone, we find satellites whose trajectory passes the UTM zone during these $5$ days, and the training samples in that UTM zone are randomly assigned across the satellites such that the number of assigned samples is proportional to the number of visits. This assignment incurs skewed distribution of labels and heterogeneity of number of samples among satellites. 
\end{itemize}

\noindent\textbf{Implementation.} 
We use DenseNet-161 \cite{huang2017densely} for the image classification task with $62$ categories. We initialize it using the pre-trained ImageNet weights~\cite{deng2009imagenet}. As batch normalization is known to be problematic in the Non-IID settings, We follow prior work~\cite{hsieh2020non} to replace batch normalization with group normalization~\cite{wu2018group}.
For preprocessing, we resize the bounding box of each images into $224\times224$ pixels with $3$ channels. 
We implement \name and three existing FL algorithms, synchronous FL, asynchronous FL, and \fedbuff for the benchmarks.
For \fedbuff, we tune the buffer size and we use the best buffer size ($M=96$) as our baseline.
For \name, GS runs aggregation scheduler per 6 hours (i.e., $I_0=24$), and set $N_\text{min}=4, N_\text{max}=8$ to reduce the search space such that $|\mathcal{R}|=5000$. 
We use a standard random forest regression to estimate the utility function $\hat{u}$ in \eqref{eq:random_search}.


\noindent\textbf{Frozen Layers.} We reduce the computational overhead at the satellites with transfer learning~\cite{TanSKZYL18}, where we freeze the lower 3 dense blocks in the DenseNet-161.

\subsection{Evaluation Results}

\begin{figure}[t!]
	\centering
	\includegraphics[height=1.5in, trim=0.05cm 0.05cm 0.1cm 0.1cm, clip]{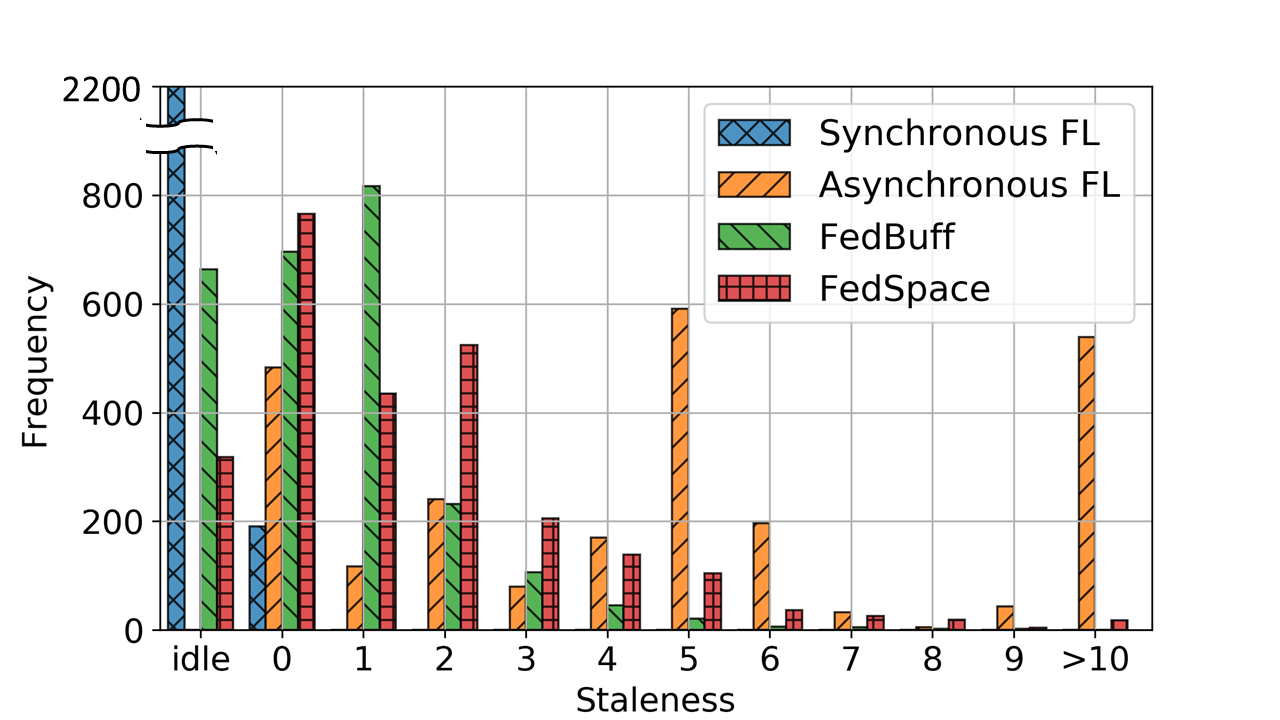}
	\vspace{-0.4cm}\caption{\footnotesize Comparison of staleness and idleness distribution among four FL algorithms. 
	\name makes the best trade-off between idleness and staleness, i.e., it has small number of idle connectivity while having large number of small value of staleness.
	Idle connectivity has no impact on training and the amount of reduced loss decreases as staleness increases. 
	}
	\label{fig:histogram_staleness}
	\vspace{-0.3cm}
\end{figure}

We measure top-1 validation accuracy using the $53,\!041$ validation samples. Figure~\ref{fig:accuracy} shows the training curve in the two dataset distribution settings. Table~\ref{tbl:target_accuracy} reports the training time to achieve the target accuracy (=$40\%$).
Figure~\ref{fig:histogram_staleness} shows a histogram of staleness and idleness distribution of the four schemes.
We make the following key observations.
\begin{itemize}[noitemsep,topsep=0pt,parsep=1pt,partopsep=2pt,leftmargin=*]
    \item In both IID and Non-IID settings, synchronous FL is unacceptably slow as more than $90\%$ of connections are idle. Most satellites spend too much time waiting for the satellites with limited connectivity.
    
    \item Asynchronous FL fails to achieve the target accuracy due to large staleness.
    
    \item \name provides substantial speedup over the \fedbuff by $0.9$ day (28.1\%) and $1.7$ day (38.6\%) in the IID and Non-IID settings, respectively. As Figure~\ref{fig:histogram_staleness} shows, \name makes better trade-off between idleness and staleness. Specifically, \name has smaller number of idle connectivity while having larger number of local updates with no staleness. 
    We observe that staleness up to 4 can provide positive impacts on model performance. 
    
    \item \name has larger performance gain in the Non-IID setting than the IID setting. 
\end{itemize}

\subsection{Discussion}\label{subsec:discussion}
Our goal in the evaluation is to demonstrate that utilizing deterministic and time-varying satellite connectivity to determine model aggregation schedule enables the efficient FL training. For simplicity, we use fMoW dataset as source dataset $\mathcal{D}^s$ to train the regression model to estimate the utility function in \eqref{eq:opt_problem}.

In the real-world setting, 
the GS can train the regression model with other source dataset such as ImageNet~\cite{deng2009imagenet} or another satellite imagery dataset~\cite{liu2017high} and then utilize transfer learning with small fraction of images downloaded from satellites. 
Another potential solution is to download the low-resolution images or to utilize another FL framework to train the regression model.

\begin{table}[t!]
\centering
\footnotesize 
\caption{Training time to achieve a target top-1 accuracy ($=40\%$) to train DenseNet-161 on fMoW dataset. Asynchronous FL fails to achieve the target accuracy. Gain represents the speed up of \name over the other schemes.}
\label{tbl:target_accuracy}
\scalebox{0.95}{
\begin{tabular}[b]{|c|c|c|c|c|c|c|c|} 
    \hline
    \multirow{2}{*}{Scheme} & \multicolumn{2}{c|}{IID} & \multicolumn{2}{c|}{Non-IID}  \\
    \cline{2-5}
       & day & gain & day & gain\\
    \hline
    Synchronous FL      & 30.3 & $13.3\times$ & 45.8 & $16.5\times$ \\
    \hline
    Asynchronous FL     & - & - & - &  - \\
    \hline
    \fedbuff  & 3.2 & $1.4\times$ & 4.4 &  $1.7\times$ \\
    \hline
    \name     & 2.3 & n/a & 2.7 &  n/a \\
    \hline
\end{tabular}
}
\vspace{-0.3cm}
\end{table}

\section{Related Works}
\label{sec:related_works}
To our knowledge, \name is the first practical FL framework running at satellites and ground stations. Previous sections discuss and evaluate the challenges in applying existing FL algorithms to this environment. We expand our discussion on related work here.

\textbf{FL at satellites.} A couple of recent works consider running FL at satellites. \citeauthor{chen2021satellite} show that FL at satellites is a feasible alternative to centralized training at ground stations~\cite{chen2021satellite}, but the study does not propose new FL algorithms. \citeauthor{razmi2022ground} also considers similar settings, but the proposed algorithm, FedSat~\cite{razmi2022ground}, only works if every satellite visits the GS exactly once per orbital period. In contrast, we formulate an optimization problem that captures the fundamental trade-offs between idleness and staleness, and we propose a general FL framework that works for any satellite networks.

\textbf{Communication-efficient FL.} It is well understood that communication is a key bottleneck in FL. Existing work mostly focuses on reducing the \emph{size} of communication using model/gradient compression (e.g., \citealt{KonecnyMYRSB16, SureshYKM17, AlistarhG0TV17, RothchildPUISB020}), and these algorithms still assume synchronous FL. These algorithms are largely orthogonal to our solution and can be combined together to reduce communication overheads.


    

\section{Conclusion}
\label{sec:conclusion}
Data collected by large satellite constellations have the potential to enable new classes of ML applications, but it is increasingly infeasible to download all the satellite data and train the ML models on the ground.
FL is a promising approach to train ML models over satellite data, but existing FL algorithms cannot address the fundamental trade-offs between idle connectivity and local model staleness. 
We formulate the optimization problem to capture the fundamental trade-offs, and we introduce an effective solution by exploiting the deterministic and time-varying connectivity among ground stations and satellites. 
We demonstrate the effectiveness of our solution with a real-world satellite imagery dataset and a satellite constellation. 
We hope that the findings and insights in this work will spur further research to design more effective FL algorithms that benefit these satellite-based ML applications.



\clearpage
\bibliography{main}
\bibliographystyle{icml2022}


\newpage
\appendix
\onecolumn


\section{Detailed Explanation on the Illustrative Example in Section \ref{subsec:exsiting_FL}}\label{app:example}

In this section, we explain the details of the illustrative example of three existing FL algorithms  depicted in Figure~\ref{fig:example_1} and Figure~\ref{fig:example_fedbuff} in Section \ref{subsec:exsiting_FL}. 
Goal of this illustrative example is to show 1) how three FL algorithms is applied among satellites and GS, and 2) the fundamental challenges of each algorithm in terms of staleness and idleness of local training. 

We consider a simple satellite constellation consisting of GS and three satellites with heterogeneous connectivity. 
In Figure~\ref{fig:example_1}, green circle at row $k\in\{1,2,3\}$ and column $i\in\{0,1,\ldots,8\}$ represents that satellite $k$ is connected to GS, i.e., $k\in\mathcal{C}_i$. 
As shown in Figure~\ref{fig:example_1}, the third satellite has limited number of connections to GS while other satellites are connected to GS four or five times. 
For each connection, blue or red arrow represents that satellite $k$ receives the global model from GS or transmits local update to GS, respectively. 
A shadow block in upper and right side of Figure~\ref{fig:example_async} shows four steps between GS and satellite $k\in\mathcal{C}_i$. 
At first, satellite $k$ sends the pair of local update $\mathbf{g}_k$ and training round index of the base global model $i_{g,k}$ to the GS. 
Second, GS stores the pair of $\mathbf{g}_k$ and $s_k$ where $s_k=i_g - i_{g,k}$ denotes staleness of $\mathbf{g}_k$, and then GS updates the global model $\mathbf{w}_i$ and global training round index $i_g$ if necessary.
Third, GS sends the pair of $\mathbf{w}_i$ and $i_g$ to satellite $k$ if it is not sent before. 
Finally, upon receiving the pair, satellite $k$ starts to train local model and stores $i_g$ as $i_{g,k}$. Now, we illustrate training procedure of three algorithms one by one.

\noindent\textbf{Synchronous FL.}
Satellite $1,2$ and $3$ receive global model and start local training at $i=0,1$ and $0$, respectively. GS downloads the local update from satellite $1,2$ at $i=4,5$ but it should wait for local update of satellite $3$ until $i=7$ to update the global model. As the global model is not updated, first two satellites remain as \emph{idle} and do nothing through connections before $i=7$.
As summarized in Table~\ref{tbl:summary_example}, synchronous FL (named Sync) has only three gradients to be aggregated while there are $8$ connections from $i=2$ to $i=8$. 
In Sync FL, satellite $3$ works as straggler which makes the connectivity of the other satellites idle, and hence significantly slow down the overall training process. 

\noindent\textbf{Asynchronous FL.}
There is no idle connection in asynchronous FL as GS updates the global model whenever it has any connection. 
On the other hand, it suffers from large staleness problem.
That is, the local gradient sent from satellite $3$ at $i=7$ is outdated as the global model is updated five times between uploading of the base global model($i=0$) and downloading of the local update ($i=7$). 
Local update with large staleness can have negative impact on the training.
As shown in Table~\ref{tbl:summary_example}, asynchronous FL (named Async) has local gradients with larger staleness even though it has no idleness. 

\noindent\textbf{\fedbuff with buffer size $M=2$.}
\fedbuff can adjust the frequency of aggregation by selecting the design parameter $M$ properly. 
GS update the global model less frequently with larger value of $M$.
We can view the synchronous FL and asynchronous FL as special case of \fedbuff with $M=1$ and $M=K$, respectively. 
In this example, by selecting $M=2$, the largest value of staleness if reduced to $2$ while there is no idle connectivity as shown in Figure~\ref{fig:example_fedbuff} and Table~\ref{tbl:summary_example}. 



\end{document}